\documentclass[10pt,twocolumn,letterpaper]{article}

\usepackage{iccv}
\usepackage{times}
\usepackage{epsfig}
\usepackage{graphicx}
\usepackage{amsmath}
\usepackage{amssymb}

\usepackage{algorithm,algorithmic}

\usepackage[toc,page]{appendix}
\usepackage[accsupp]{axessibility}

\usepackage[pagebackref=true,breaklinks=true,letterpaper=true,colorlinks,bookmarks=false]{hyperref}

\iccvfinalcopy 

\def\iccvPaperID{25} 

\ificcvfinal\fi

\begin{document}

\title{DetOFA: Efficient Training of Once-for-All Networks \\ for Object Detection using Path Filter}

\author{
Yuiko Sakuma \thanks {Corresponding author} \qquad 
Masato Ishii \qquad 
Takuya Narihira \\
Sony Group Corporation, Tokyo, Japan \\
{\tt\small \{Yuiko.Sakuma, Masato.A.Ishii, Takuya.Narihira\}@sony.com}
}

\maketitle
\ificcvfinal\thispagestyle{empty}\fi

\begin{abstract}
   We address the challenge of training a large supernet for the object detection task, using a relatively small amount of training data. Specifically, we propose an efficient supernet-based neural architecture search (NAS) method that uses search space pruning. The search space defined by the supernet is pruned by removing candidate models that are predicted to perform poorly. To effectively remove the candidates over a wide range of resource constraints, we particularly design a performance predictor for supernet, called path filter, which is conditioned by resource constraints and can accurately predict the relative performance of the models that satisfy similar resource constraints. Hence, supernet training is more focused on the best-performing candidates. Our path filter handles prediction for paths with different resource budgets.  Compared to once-for-all, our proposed method reduces the computational cost of the optimal network architecture by 30\% and 63\%, while yielding better accuracy-floating point operations Pareto front (0.85 and 0.45 points of improvement on average precision for Pascal VOC and COCO, respectively).
\end{abstract}

\section{Introduction} \label{sec:introduction}

Object detection is one of the fundamental computer vision tasks, which has been widely used in real-life applications, such as on smartphones, for surveillance, and autonomous driving. In the past decade, deep neural network-based methods have become the state-of-the-art method, yielding very high performance \cite{ren2015faster, lin2017focal}. The recently proposed CenterNet \cite{zhou2019objects} is a simple anchor-free, one-stage method that has high computational efficiency. 

\begin{figure}[t]
\begin{center}
   \includegraphics[width=1.0\linewidth]{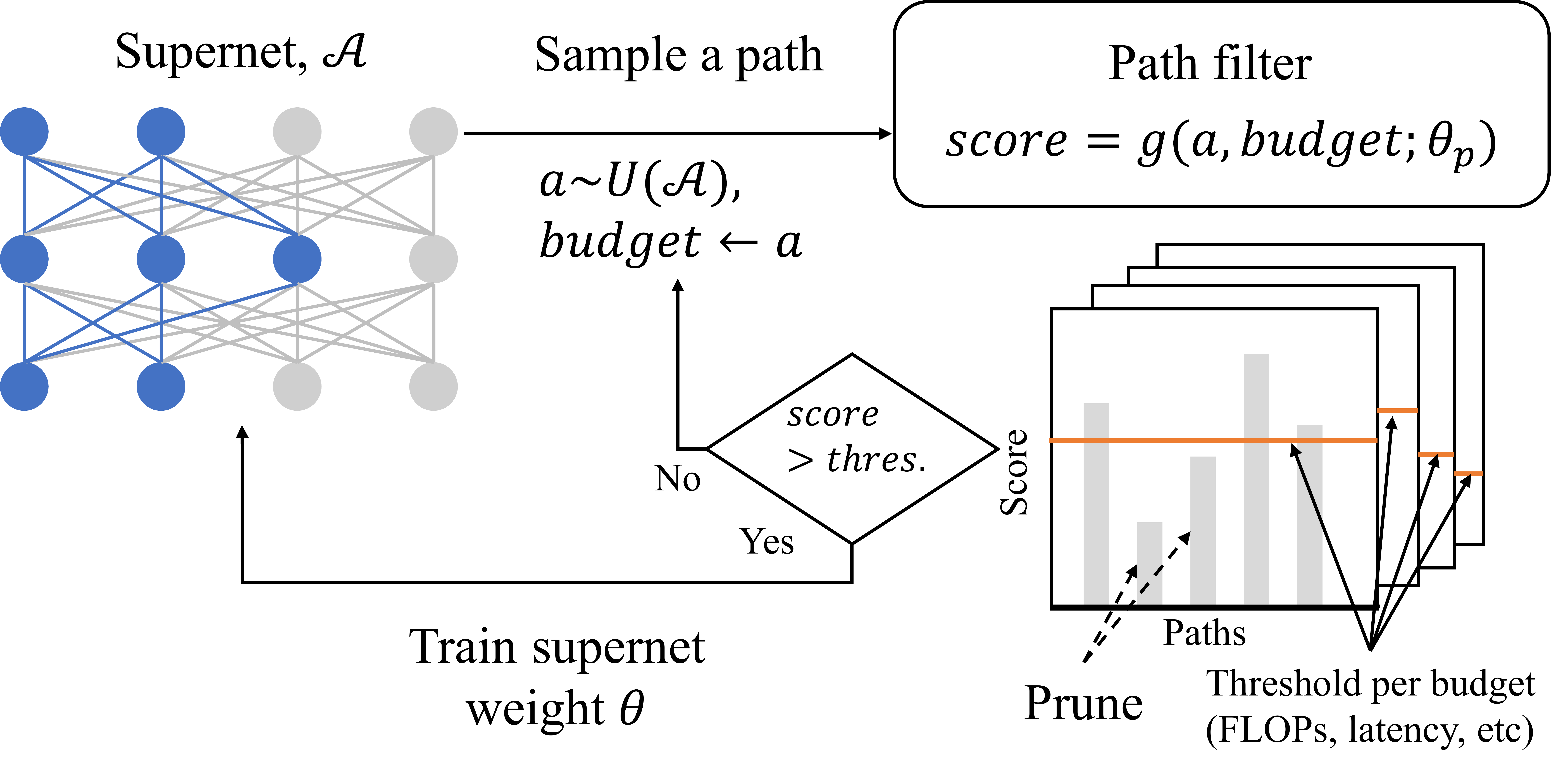}
\end{center}
   \caption{Our proposed search space pruning method using a path filter. A path is sampled from the supernet, $\mathcal{A}$. The path filter predicts the path score, the relative performance of a path within the budget, from the architecture and budget information. The weak-performing (e.g., low accuracy) paths are evenly pruned from the search space considering the budget.Thus, the supernet training is more focused on good-performing paths.
   }
\label{fig:overview}
\end{figure}

As the efficiency requirements for detection models get more pronounced for various applications \cite{chen2019detnas, jiang2020sp}, the challenge of efficient deployment arises. In particular, models often are deployed to a diverse set of hardware platforms \cite{bu2021gaia} and, hence, have to meet different resource constraints. For instance, the latest smartphones and edge devices such as surveillance cameras have different processing speeds and memory capacities. Designing performant model architectures that meet resource constraints across different devices is a laborious task and requires high computation costs. Recently proposed supernet-based neural architecture search (NAS) methods \cite{cai2019once, yu2020bignas, chu2021fairnas, guo2020single, wang2020hat} decouple the model training and search and achieve remarkable search efficiency and final model performance. Specifically, supernet-based NAS methods are efficient because the supernet is trained only once and can be scaled to fit any computational budget without any retraining, later. Up to now, supernet-based NAS has mostly been used to optimize network architectures for classification tasks, where large training sets are available, and has only recently been adopted for some object detection tasks \cite{bu2021gaia}.

One of the major challenges in supernet-based NAS for object detection is how to efficiently train the supernet with limited training data \cite{girshick2015region}. For instance, while commonly used ILSVRC2012 (ImageNet) \cite{deng2009imagenet} contains 1.2 million images with 1000 classes, Pascal visual object classes (Pascal VOC) \cite{everingham2009pascal} contains 16,000 images with 20 classes, which is only 1\% in data size of ImageNet. On the other hand, the search space of supernet-based NAS is large; for example, the search space of Once-for-All (OFA) \cite{cai2019once}, one of the most referenced supernet-based NAS, contains $10^{19}$ architectures. Supernet training requires training multiple sub-networks (e.g., paths) in the search space simultaneously. When a small dataset is available, it is prone to under-fit, which leads to a noticeable degradation in the model performance.

We propose a new training method for supernet-based NAS, which is suitable for tasks where only a small amount of training data is available (Figure \ref{fig:overview}). Although we introduce our method in the example of object detection, it is general and can be used to train supernets for any task. Our efficient NAS method uses search space pruning via a path filter. We propose to prune the supernet after pre-training to effectively reduce the number of candidate paths in the search space. Specifically, we use a path filter to discard weak-performing operations and paths and focus more on good ones. Here, paths are the candidate architectures, which are often called sub-networks contained in the supernet. Operations are the layer configurations of a path. The path filter is trained with a ranking loss, using path configuration/performance pairs, that are collected on the validation data. Further, the path filter is pre-trained by sampling paths from the pre-trained supernet and transferred to the target task as well. Because our method effectively prunes the search space, the supernet can be trained well with relatively small object detection data. 

To our knowledge, our work is the first to propose a path filter that is specifically designed for supernet, which is conditioned by constraints and can be used to evenly prune the search space along the resource constraint for training good-performing paths in a one-shot manner. Moreover, the supernet architecture for object detection has not been explored well. Our target includes low-memory edge devices with small FLOPs constraints, thus, we chose CenterNet \cite{zhou2019objects} because of (1) its widespread usage in real applications and (2) its efficiency in FLOPs. 

In experiments, we demonstrate that our method yields better-performing supernets for object detection tasks. Our method achieves 30\% and 63\% reduction in computation cost while 0.85 and 0.45 points improvements in the average precision (AP) over different floating-point operations (FLOPs) constraints for Pascal VOC and Microsoft COCO (COCO) \cite{lin2014microsoft}, respectively, compared to OFA. 


\section{Related works}

\subsection{Object detection}
Popular object detection methods are discussed in the supplementary material \ref{sec:related works on object detection}. To summarize, among two-stage detectors \cite{girshick2015fast, ren2015faster} and one-stage detectors \cite{redmon2016you, lin2017focal, bochkovskiy2020yolov4}, CenterNet \cite{zhou2019objects} and FCOS \cite{tian2019fcos} are simple anchor-free, one-stage methods. CenterNet models an object as the center point of its bounding box. First, it finds center points, while other object properties like size and orientation are regressed after. The model comprises a convolution neural network (CNN) backbone, up-convolutional layers, and three heads. Each head predicts a keypoint heatmap and a set of local offsets and object sizes. CenterNet is a general method that can be used for other vision tasks such as human pose estimation and 3D object detection, as well.


\subsection{NAS}
Recently, NAS has been studied to automatically design good-performing neural network architectures under different resource budgets. NAS methods, including evolutionary search \cite{dai2020fbnetv3, dai2019chamnet}, reinforcement learning \cite{tan2019mnasnet}, or one-shot method \cite{liu2018darts}, identify the optimal network architecture for a given resource budget from a set of candidate architectures (the search space). This typically involves training and search steps.
Recently, supernet-based methods have been proposed \cite{cai2019once, yu2020bignas, chu2021fairnas, guo2020single, wang2020hat}, which decouple the model training and search. The search space of supernet-based NAS contains multiple paths. OFA \cite{cai2019once} is a memory-efficient supernet-based NAS that shares the supernet weights. For training the large supernet with the many shared weights, OFA adapts the progressive shrinking strategy, which gradually trains the large paths to small ones.

\textbf{NAS for object detection}:
DetNAS \cite{chen2019detnas}, SP-NAS \cite{jiang2020sp}, and DetNAS \cite{chen2019detnas} aim to optimize the backbone architecture. However, they only yield one single optimal architecture that meets a single hardware constraint. A problem is, that because the NAS search space contains many weights and because object detection datasets are small, the supernet easily overfits. Transfer learning is a common approach to address the limitation of a small target dataset. Object detection models are for example often pre-trained with classification datasets, i.e., DetNAS uses the backbone initialized with ImageNet. GAIA \cite{bu2021gaia} pre-trains the supernet using a huge data pool with a unified label space from multiple data sources. Then, the downstream fine-tuning is performed on the target dataset. Although GAIA provides a powerful pre-training model, the data unification process requires massive labeling efforts. While GAIA emphasizes the effectiveness of the pre-training dataset, we study the effectiveness of different pre-training methods.


\subsection{NAS with search space pruning}

\subsubsection{Search space pruning for single constraint NAS} 

Search space pruning techniques are categorized into path \cite{you2020greedynas, su2021prioritized} and operation levels \cite{hu2020angle, shen2020bs}. GreedyNAS \cite{you2020greedynas} produces a path candidate pool to store good paths and samples from it under an exploration-exploitation strategy. MCTNAS \cite{su2021prioritized} proposes a sampling strategy based on the Monte Carlo tree search. ABS \cite{hu2020angle} proposes an angle-based metric that measures the similarity between initialized and trained weights to predict the generalization ability of paths. BS-NAS \cite{shen2020bs} proposes a channel-level importance metric.

While operation-level pruning is more efficient because several paths are pruned by pruning one operation, selecting operations to be pruned is not trivial, i.e., evaluating the impact of pruning specific operations on the paths' performance or efficiency is not an easy task. However, it is more straightforward for paths such that validation loss or accuracy can be easily estimated. GreedyNASv2 \cite{huang2022greedynasv2} considers both path and operation level pruning. A path filter, trained to predict weak paths, is used to prune them from the search space. They argue that identifying weak paths is more reliable than identifying good ones. The search space is further pruned by removing operations; the operations with similar path filter embedding are merged into the ones with smaller FLOPs. Similar to GreedyNASv2, our method uses a path filter to identify the paths and operations to be pruned. However, our goal is to train a supernet from which we can extract paths that meet different resource constraints. In comparison, GreedyNASv2 only yields a single network architecture that is optimal for a given resource constraint. In particular, GreedyNASv2 must be run from scratch if the resource constraints change. Moreover, while GreedyNASv2 uses a path filter that predicts the binary label (that is, weak path or not) of each path, our path filter learns a path ranking. Thus, our path filter is more flexible, i.e., once our path filter is trained, different pruning ratios can be applied. In comparison, the path filter proposed for GreedyNASv2 needs to be retrained for each pruning ratio.


\subsubsection{Search space pruning for supernet-based NAS}

While NAS typically employs uniform sampling for the supernet training \cite{cai2019once}, iNAS \cite{gu2021inas} proposes a sampling method with latency grouping, which groups the operations with the same latency across layers, to train a supernet for salient object detection. Although this method limits the operation combinations and, hence prunes the search space, using the latency as a pruning criterion might not be optimal. For instance, the candidate paths do not include the ones with low latency operations in the input side layers and high latency operations in the output side. AttentiveNAS \cite{wang2021attentivenas} focuses on the best and worst performing paths for supernet training. However, they only focus on path-level pruning. CompOFA \cite{sahni2021compofa} prunes the search space by coupling the operations (Table \ref{tab:search space}). It halves the computation time without performance degradation compared to OFA's progressive shrinking. However, their method is heuristic because the operation combination is determined without consideration of the path performance. Ours is more optimal and general because the operation candidates are determined based on the path score (that is, performance) and FLOPs. Further, no prior works in this category consider joint path and operation-level pruning.


\section{Proposed method}

As discussed in Section \ref{sec:introduction}, we propose an efficient supernet-based NAS method that uses search space pruning via a path filter, and apply it to object detection. Let $f(x, a; \theta)$ be a supernet with input $x$, path $a$, and weights $\theta$ that is initialized by a pre-training task and  $y$ be the output. For the CenterNet, the output $y$ is the key-point heatmap, local offset, and object size. We propose to prune the search space while fine-tuning. A path filter $g(a; \theta^*_p)$ that is trained to rank paths in the search space is used to prune the search space. The output of the path filter is the path score. We prune the operations and paths that are predicted to perform badly. Specifically, the supernet is fine-tuned for the target task by computing the optimized shared weights $\theta^*$, according to;
\begin{equation}
    \label{eq:supernet fine-tuning}
    \begin{split}
        &\underset{\theta}{\mathrm{min}} 
        \mathbb{E}_{a\sim U(\mathcal{\Tilde{A}})}
        \left[\frac{1}{N} \sum^N_i 
        \mathcal{L}(f(x_i, a; \theta), y_i) \right] \\
        &(x_i, y_i) \in \mathcal{D}_{\mathrm{trn}}
    \end{split}
\end{equation}
where $\Tilde{\mathcal{A}}$ ($|\Tilde{\mathcal{A}}| \ll |A|$) and $\mathcal{D}_{\mathrm{trn}}$ are the search space pruned from the original one $\mathcal{A}$ and training dataset, respectively.

After training the supernet, the optimal path $a^*$ is searched that yields the best path score on the target task for a given FLOPs budget as follows:
\begin{equation}
    \label{eq:search}
    \begin{split}
        &a^* = 
        \arg \underset{a \in \Tilde{A}}{\mathrm{max}} \, g(a; \theta^*_p) \\
        &s.t. \: \mathrm{FLOPs}(a) \leq \tau
     \end{split}
\end{equation}
where $\mathrm{FLOPs}(a)$, and $\tau$ are the FLOPs of path $a$, and FLOPs budget, respectively. Although FLOPs are used in the experiment, other resource constraints such as latency and memory size can also be used. The details of path filter and search space pruning are provided in Section \ref{sec:path filter} and \ref{sec:search space pruning}, respectively. The details of resource constraint search are provided in Section \ref{sec:resource-constraint search}. 


\begin{figure}[t]
\begin{center}
\includegraphics[width=0.8\linewidth]{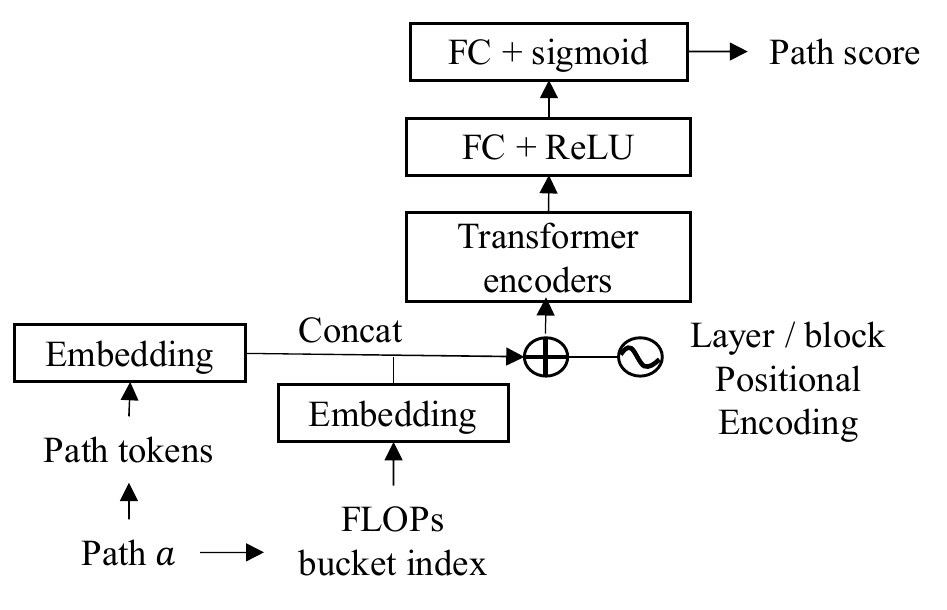}
\end{center}
   \caption{The architecture of the path filter. We use a transformer encoder to predict performance scores from tokenized paths.}
\label{fig:path filter}
\end{figure}

\begin{figure}[t]
\begin{center}
\includegraphics[width=0.8\linewidth]{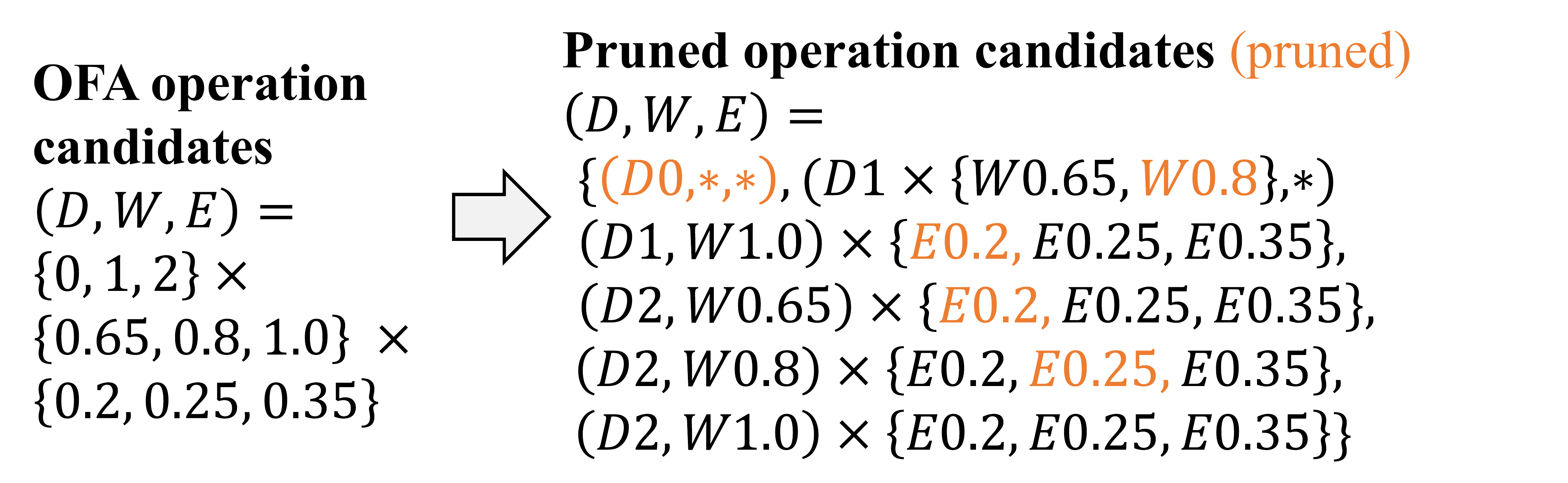}
\end{center}
   \caption{Example of operation pruning. $D$, $W$, and $E$ denote the bottleneck depth, channel width expand ratio, and channel expand ratios. The operation candidates are pruned according to the predicted path scores and FLOPs, thus reducing the possible combinations of them.}
\label{fig:operation pruning}
\end{figure}

\subsection{FLOPs-bounded path filter} \label{sec:path filter}
Given the architecture configuration of a path, the path filter predicts the relative performance of the path among all path candidates in the search space. It is used to specify weak-performing paths and operations to be pruned from the search space. Details for path filter design and training are provided below.

\subsubsection{Path filter design} \label{sec:path filter design}

Figure \ref{fig:path filter} shows the architecture of our path filter.
A transformer encoder architecture \cite{vaswani2017attention} is chosen because it yields state-of-the-art performance for many tasks and a high computation efficiency due to the parallelization ability. The architecture configuration and the associated number of FLOPs are represented by path tokens and a FLOPs bucket index, respectively. First, both are converted into feature vectors via a learned embedding and then concatenated. The embeddings are enriched with positional encodings of the architecture configuration and fed to the transformer encoder. The final path score is computed, using two fully connected (FC) layers with sigmoid activation.

\textbf{Tokenization}:
The architecture configuration of a path $a \sim U(\mathcal{A})$ is encoded by using layer-wise tokenization with positional encoding. Here, we use OFA ResNet50 \cite{mit-han-lab} as an example to explain how tokenization works. The configuration of a path in OFA ResNet50 has a fixed layer connectivity and is completely defined by the depth ($D$) of a block, the channel width ($W$), and the channel expansion ratio ($E$). Here, $W$ and $E$ define the channel size of a whole block and within a bottleneck in a block, respectively. The OFA network consists of bottleneck blocks with several layers; operation configurations are either block-wise (that is, D and W) or layer-wise (that is, E). We propose to tokenize on a per-layer basis; the block-wise configurations are simply repeated for each block. Each layer can be represented by a token consisting of W, E, and skip connection (SC) derived from D. Here, an example of tokenization for a bottleneck block with a maximum depth of four (that is, four layers) is presented. When $($D$, $W$, $E$) = (0, 1.0, (0.25, 0.35))$, the token would be "W1.0\_E0.25 W1.0\_E0.35 SC SC" where SC denotes the skip connection, such that the bottleneck with $D=0$ has only two layers.

As detailed in Section \ref{sec:path filter training}, since the path score is only used to compare paths within the same bucket, FLOPs information is added as the input. The FLOPs bucket is determined from the minimum and maximum FLOPs among all paths in $\mathcal{\Tilde{A}}$ by binning values into discrete intervals of equal widths. In the experiment, we set the number of FLOPs buckets as five. The required number of FLOPs to compute the output of a path is represented by the corresponding bucket index. This tokenization scheme can be generalized to other OFA supernets and search spaces.
However, the positional encoding is not as simple in the case of more complex layer connectivity.

\textbf{Layer and block positional encoding}:
Since the self-attention mechanism used in transformers is agnostic to the position of the data in a sequence, the layerwise encodings are enriched with positional encodings. More specifically, we use standard cosine/sine-based positional encodings that are defined as
\begin{equation}
    \label{eq:layer positional encoding}
    \begin{split}
         P(l, 2i)=\mathrm{sin}(\left. l \middle/ 10000^{\left. 2i \middle/ \mathrm{max}\_{l} \right.} \right.) \\
         P(l, 2i+1)=\mathrm{cos}(\left. l \middle/ 10000^{\left. 2i \middle/ \mathrm{max}\_{l} \right.} \right.)
    \end{split}
\end{equation}
where $l$ and $i$ denote the layer index and embedding dimension, respectively. $max\_l$ is the maximum layer index in $\Tilde{\mathcal{A}}$. Further, the block index of the bottleneck is encoded. As discussed above, for OFA supernets, some operations are block-wise, while others are layer-wise. Thus, block-level encoding adds extra information to the path filter. The formulation is similar to the layer-wise positional encoding (Equation \ref{eq:layer positional encoding}) but replaces $l$ by the block index.


\begin{figure}[t]
\begin{center}
   \includegraphics[width=1.0\linewidth]{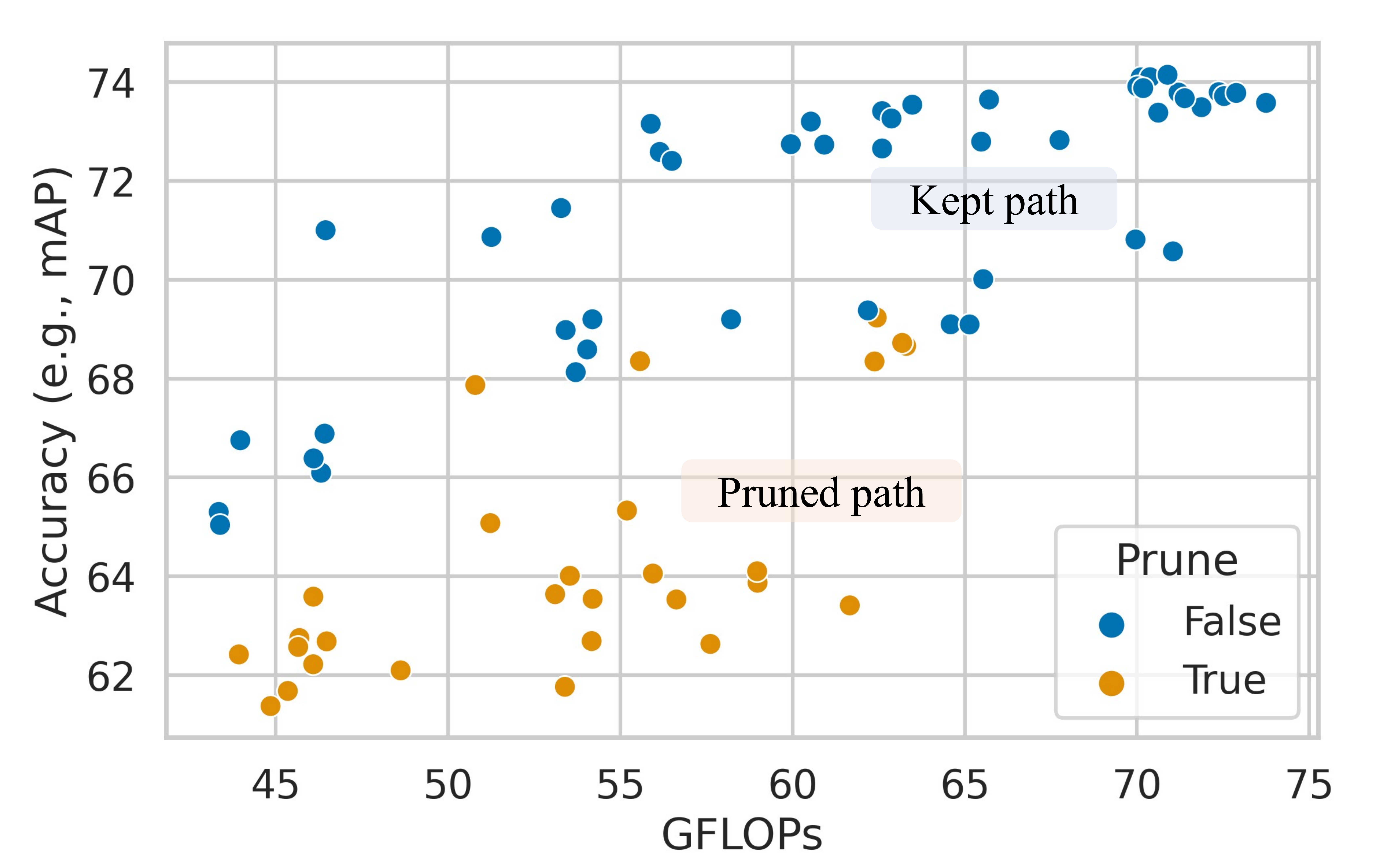}
\end{center}
   \caption{Path pruning via path filter. The weak-performing paths are pruned during supernet training.}
\label{fig:path pruning}
\end{figure}

\subsubsection{Path filter training} \label{sec:path filter training}

We aim to predict weak-performing (e.g., low accuracy) paths and operations for any pruning ratio with good precision. Our path filter learns a full ranking of the path performance and can be used for different pruning ratios. while GreedyNASv2 can only be used for a fixed pruning ratio. Specifically, the output of our path filter is a path score that indicates the relative performance of the path; higher scores mean lower validation loss. Once the path ranking is learned, it can be used to prune weak-performing paths from the supernet, using any pruning ratio. 
The learning-to-rank problem is equivalent to the area under the ROC curve (AUC) maximization \cite{yuan2021large, yang2022auc}. The pairwise surrogate loss is often used for AUC maximization. Further, we propose the \textbf{FLOPs-bounded loss}. Since for supernet-based NAS, the ranking among paths with similar FLOPs matters, the loss is calculated only for path pairs that fall into the same FLOPs bucket, i.e.
\begin{equation}
    \label{eq:path filter optimization}
    \begin{split}
        g_{aa'} &:= g(a; \theta^*_p) - g(a'; \theta^*_p) \\
        \theta_p^* &= \underset{\theta_p}{\mathrm{min}} 
        \mathbb{E}_{a, a' \sim U(\mathcal{A})}
        \left[\mathcal{L}_p(g_{aa'}, t_{aa'}) \right] \\
        t_{aa'} &= 
        \begin{cases}
            1, \text{if} \frac{1}{N_p}
            \sum^N_i \mathcal{L}(f(x_i, a; \theta^*), y_i) \leq \\
            \hspace{20pt} \frac{1}{N_p} \sum^N_i \mathcal{L}(f(x_i, a'; \theta^*), y_i) \\
            \hspace{20pt} \text{and} \: \mathrm{FLOPs}(a') \leq \tau_{\mathrm{FLOPs}(a)} \\
            0, \text{otherwise} 
        \end{cases} \\
        &(x_i, y_i) \in \mathcal{D}_{\mathrm{val}} \\
    \end{split}
\end{equation}
The squared hinge loss 
\begin{equation}
    \label{eq:squared hinge loss}
    \begin{split}
        \mathcal{L}_p(g_{aa'}, t_{aa'}) = max(0, 1 - t_{aa'} \cdot g_{aa'})^2
    \end{split}
\end{equation}
is chosen, because it is one of the most commonly used pairwise ranking losses \cite{khalid2019scalable, yan2003optimizing}.
For training the path filter, $m$ paths are sampled from the supernet for each FLOPs bucket in order to construct a training dataset. Then, the dataset is split into 80\% training and 20\% validation samples. The path filter with the best validation loss is finally used for search space pruning. The path filter is pre-trained as well. After transferring the path filter to the target task, it is fine-tuned for some warm-up epochs.


\begin{table*}
\begin{center}
\begin{tabular}{|l|c|c|}
\hline
Method & Search space & Size \\
\hline\hline
OFA & 
$($D$, $W$, $E$) = \{\{0, 1, 2\} \times \{0.65, 0.8, 1.0\} \times \{0.2, 0.25, 0.35\}\}$ &
\begin{tabular}{c}
$((3^2 + 3^2) \times (3^3 + 3^3) \times (3^4 + 3^4))^4$ \\
$\approx 10^{20}$ 
\end{tabular} \\
CompOFA*     & 
$($D$, $W$, $E$) = \{(0, 0.65), (1, 0.8), (2, 1.0)\} \times \{0.2, 0.25, 0.35\}$  &
$(3^2 + 3^3 + 3^4) \approx 10^8 $ \\
\hline
\end{tabular}
\end{center}
\caption{The search spaces used in our experiments. $D$, $W$, and $E$ denote the bottleneck depth, channel width expand ratio, and channel expand ratios.}
\label{tab:search space}
\end{table*}

\subsection{Search space pruning via path filter} \label{sec:search space pruning}
First, operations are pruned by using the path score.
Then, during supernet fine-tuning, paths are pruned.

\textbf{Operation pruning} (Figure \ref{fig:operation pruning}): 
To evaluate the operation candidates by using the path filter that is trained to evaluate each path, $n$ paths are uniformly sampled. Then, the candidate operation is inserted in order to measure its performance for a given combination of $D$, $W$, and $E$. The average of $n$ path scores is used as the operation score. This gives the score and FLOPs information of candidate operations for all layers as depicted in Figure \ref{fig:operation pruning}.
Three pruning strategies are proposed as follows:
\begin{itemize}
    \item FLOPs: For each FLOPs bucket, $r_{\mathrm{op}}$\% operations are pruned with uniform distribution.
    \item FLOPs \& score (per bucket): For each FLOPs bucket, worst $r_{\mathrm{op}}$\% operations are pruned.
    \item FLOPS \& score (all): For all FLOPs buckets, worst $r_{\mathrm{op1}}$\% operations are pruned. Then, for each FLOPs bucket, worst $r_{\mathrm{op2}}$ operations are pruned.
\end{itemize}

\textbf{Path pruning}: 
During supernet fine-tuning, $r_{\mathrm{path}}$ weak performing paths are pruned for each FLOPs bucket. Specifically, sampling of pruned paths is skipped and the supernet is only trained for the kept paths. Figure \ref{fig:path pruning} presents the pruned and kept paths, evaluated by path filter when 15 paths are uniformly sampled from the fine-tuned supernet. Algorithm \ref{alg:supernet training} summarizes our proposed supernet training method with the path filter. For path pruning, when a path $a$ is sampled (Line 10), it is evaluated by the path filter. If it is larger than the threshold, the supernet is trained for $a$, and if not, a new path is sampled (Line 12).

\begin{algorithm}[t]
\caption{Supernet training with search space pruning using a path filter}
\label{alg:supernet training}
\begin{algorithmic}[1]
    \REQUIRE {
    Pre-trained supernet weights $\theta$,
    pre-trained path filter $g(a; \theta_p)$,
    training dataset $\mathcal{D}_{\mathrm{trn}}$, 
    search space $\mathcal{A}$,
    operation pruning ratio $r_{\mathrm{op}}$, path pruning ratio $r_{\mathrm{path}}$}, Score threshold for path $a$ $\delta_{\mathrm{FLOPs}(a)}$,
    \STATE{Initialize $\Tilde{\mathcal{A}}$ with $\mathcal{A}$}
    \STATE{Supernet warm-up training}
    \STATE{Uniformly sample $m$ paths for each FLOPs bucket and validate}
    \STATE{Fine-tune path filter to $g(a; \theta^*_p)$}
    \STATE{Prune $r_{\mathrm{op}}$\% operations from $\Tilde{\mathcal{A}}$ via path filter}
    \STATE{Calculate path score threshold $\delta_{\mathrm{FLOPs}(a)}$ with pruning ratio $r_{\mathrm{path}}$\% for each FLOPs bucket}
    \FOR{$epoch = 1, ..., \mathrm{max\_epoch}$}
        \FOR{$i = 1, ..., \mathrm{max\_iter}$}
            \STATE{Sample $(x_i, y_i) \in \mathcal{D}_{\mathrm{trn}}$}
            \STATE{$a \sim U(\Tilde{\mathcal{A}})$ (uniform sampling of a path)}
            \WHILE{$g(a; \theta^*_p) < \delta_{\mathrm{FLOPs}(a)}$}
                \STATE{$a \sim U(\Tilde{\mathcal{A}})$}
            \ENDWHILE
            \STATE Update $\theta$ via gradient descent (Equation \ref{eq:supernet fine-tuning}) 
        \ENDFOR
    \ENDFOR
\end{algorithmic}
\end{algorithm}



\subsection{Resource-constraint search} \label{sec:resource-constraint search}

Similar to OFA \cite{cai2019once}, we adopt the evolutionary search \cite{real2019regularized} algorithm to search architectures for the given resource constraint. Because evaluating the actual performance of each path during the search phase is expensive, the path score is used as the proxy as done by OFA \cite{cai2019once}. OFA for example uses an accuracy predictor that regresses the exact validation accuracy. The path filter can also be used for standard supernet-based NAS, and not only for our proposed method. We used the path filter as the accuracy predictor in the search phase. Note, that before the search the path filter is re-trained. Similar to the supernet warm-up (Line 2 in Algorithm \ref{alg:supernet training}), $m$ paths are uniformly sampled for each FLOPs bucket to obtain training data.


\section{Experiments}

\begin{figure}[t]
\begin{center}
   \includegraphics[width=1.0\linewidth]{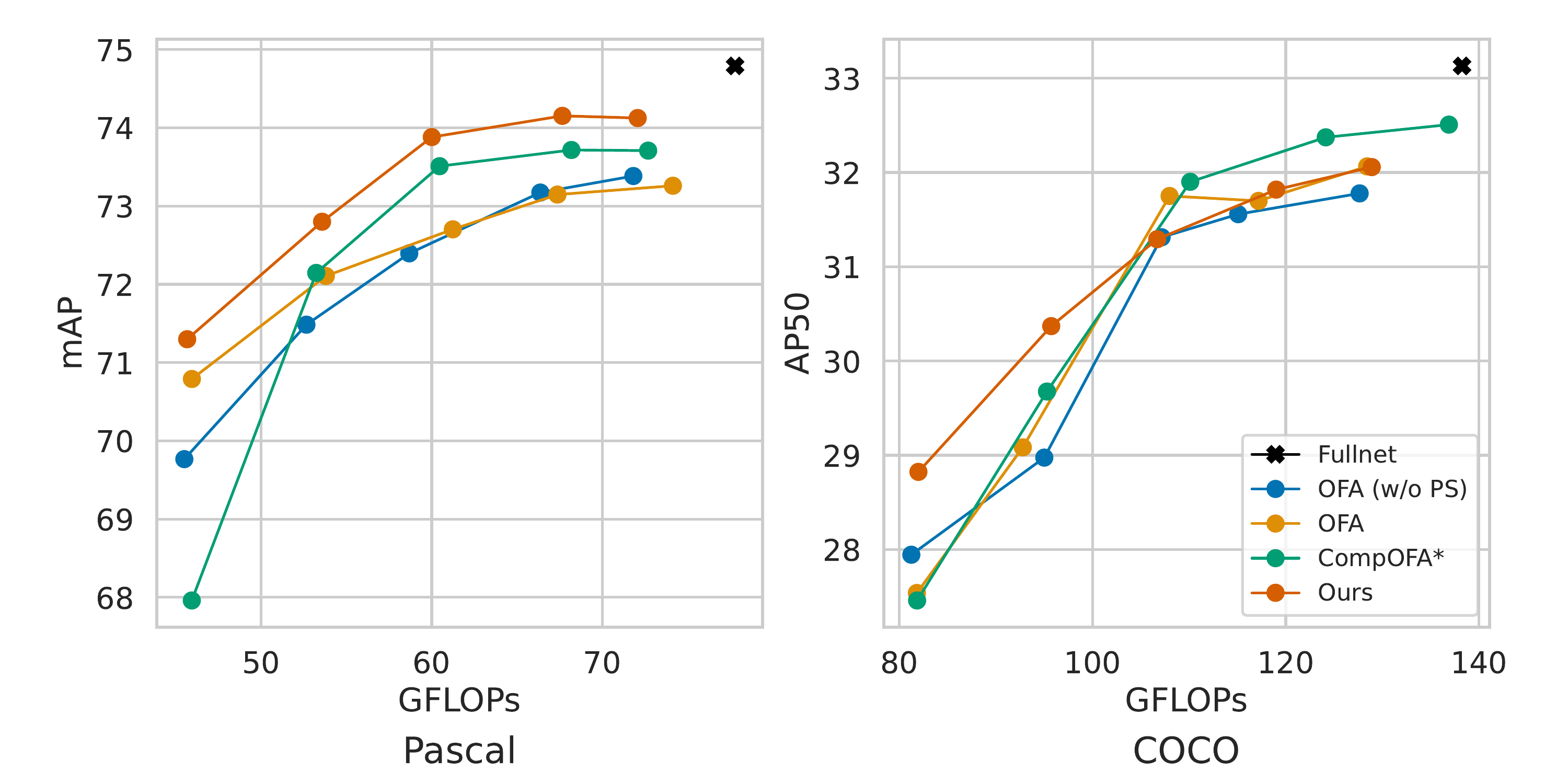}
\end{center}
   \caption{The performance of the optimal architecture $a^*$, for object detection on the Pascal VOC (left) and the COCO (right) dataset. Our method outperforms OFA (w/o progressive shrinking, PS) across all given FLOP bounds. CompOFA* performs well only for larger FLOPs. Fullnet denotes the largest path after fullnet training (Line1 in Algorithm \ref{alg:supernet training}).}
\label{fig:evolution search}
\end{figure}

\begin{figure}[t]
\begin{center}
   \includegraphics[width=1.0\linewidth]{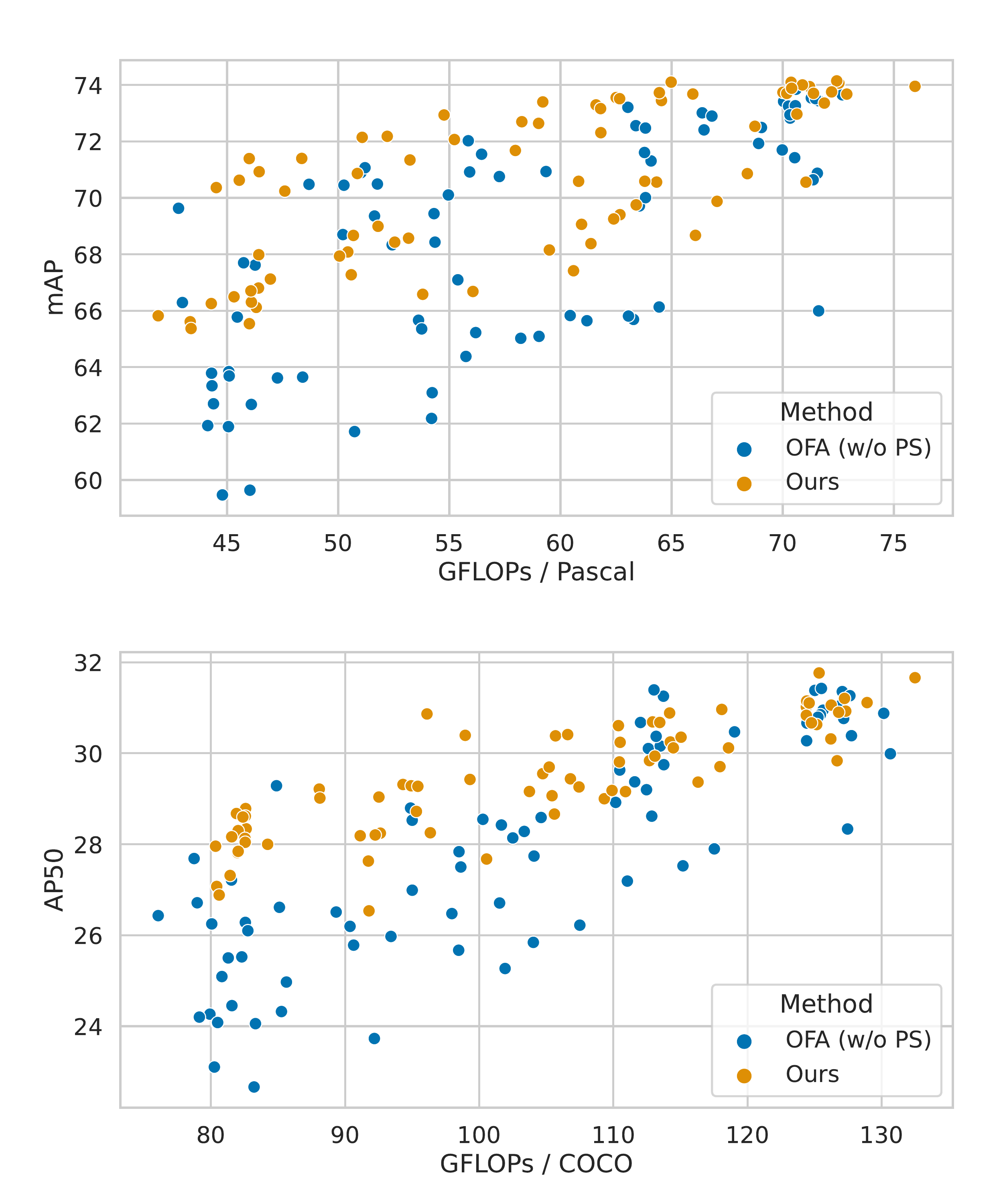}
\end{center}
   \caption{The validation performance of 15 uniformly sampled paths across $5$ FLOPs buckets on the Pascal VOC (top) and the COCO (bottom) dataset. Our proposed method yields paths with a better average performance.}
\label{fig:eval search space}
\end{figure}

\subsection{Experimental setup}
The experiments are implemented, using the NNablaNAS framework \cite{nnablanas}. Experiments on Pascal VOC and COCO were performed on NVIDIA V100 and A100 GPUs, respectively. We use the same search space as \cite{mit-han-lab}, which yields the OFA-ResNet50 supernet summarized in Table \ref{tab:search space}.

\textbf{Path filter}: The path filter uses an embedding layer with 128 dimensions, three transformer encoders (with four-head self-attention and an FC layer with 128 dimensions), and two final FC layers with 128 dimensions. The path filter was pre-trained with path configuration/performance pairs obtained by sampling 10000 paths in total from the supernet trained on ImageNet. Path filter fine-tuning was conducted by sampling 500 paths after fine-tuning the supernet on the detection dataset.

\textbf{Transfer learning details}:
The CenterNet backbone is transferred from the supernet pre-trained by the classification task.
A max-pooling layer and classification head are replaced by task-specific layers, which are up-convolution layers and detection heads.

\textbf{Supernet training and search details}:
The supernet was trained for 5 and 10 warm-up epochs for Pascal VOC and COCO, respectively. Then, it was fine-tuned for 65 and 60 epochs, respectively. Training details are described in the supplementary materials \ref{sec:training details}. The number of generations in the evolutionary search was 500.


\subsection{Comparison with prior NAS approaches}

\begin{table*}
\begin{center}
\begin{tabular}{|c|c|c|c|c|c|}
\hline
FLOPs-bounded loss & Budget condition & Block positional encoding & 
Accuracy & Precision & Recall \\
\hline\hline
--         & \checkmark & \checkmark & 0.790 & 0.590 & 0.174 \\
\checkmark & --         & \checkmark & 0.900 & 0.830 & 0.209 \\
\checkmark & \checkmark & --         & 0.940 & 0.850 & 0.190 \\
\checkmark & \checkmark & \checkmark & 0.930 & 0.870 & 0.224 \\
\hline
\end{tabular}
\end{center}
\caption{Evaluation of the path filter for detecting the weakest 25\% of the paths. Results with the best precision are presented.}
\label{tab:path filter performance}
\end{table*}

The proposed method is compared to the prior supernet-based NAS methods, OFA and CompOFA. Note that the ResNet50-based supernet architecture was originally not proposed for CompOFA. However, to be comparable, we define a similar supernet proposed in the CompOFA paper\cite{sahni2021compofa} as given in Table \ref{tab:search space} and denote it as CompOFA*. In this experiment, OFA with and without progressive shrinking (denoted as w/ and w/o PS, respectively) were compared. OFA (w/ PS) were trained by $\times 3$ epochs than the proposed method. OFA (w/o PS) and CompOFA were trained for the same epochs (that is, warm-up + fine-tuning epochs) as the proposed method for a fair comparison. The detailed training schedules are presented in supplementary materials \ref{sec:training details}. Here, the operation pruning method is "FLOPS \& score (all)", removing 10\% of worst performing operations overall and 30\% of them per each FLOPs range. The path pruning ratio was 25\%.

Figure \ref{fig:evolution search} presents the evolution search results on the average of three runs. Transfer learning improves the accuracy by 6.55 and 1.85 points on average for Pascal VOC and COCO, respectively. Our method achieves better performance over OFA (w/ and w/o PS) over the different FLOPs. For instance, compared to OFA (w/o PS), the proposed method outperforms by 0.85 and 0.45 points on average for Pascal VOC and COCO, respectively. Specifically, our method performs better than OFA (w/ and w/o PS). CompOFA* performs well for large FLOPs, but weaker for smaller paths. This result presents the limitation of hand-crafted search space pruning. Paths with larger depths and smaller channels on the output side performed better for small paths. This corresponds to the intuition that detection models have to keep rich feature information in the backbone to be processed in up-convolutional and head layers. CompOFA* removes these paths by coupling small depths and channels together. This result suggests that handcrafted search space in CompOFA is well-tuned only for a specific task and a search space, while our method is more general and potentially works well for arbitrary search spaces and tasks.

\begin{table}
\begin{center}
\begin{tabular}{|l|c|c|}
\hline
Stage & Pascal VOC &  COCO \\
\hline\hline
OFA (w/ PS)  &  1.150 & 14.41 \\
OFA (w/o PS) &  0.383 & 4.804 \\
CompOFA*     &  0.383 & 4.804 \\
Ours         &  0.800 (0.414) & 5.260 (0.456) \\
\hline
\end{tabular}
\end{center}
\caption{Computation cost for sampling paths and supernet training in GPU days. The values in brackets denote computation time for path sampling for path filter fine-tuning.}
\label{tab:training cost}
\end{table}

Table \ref{tab:training cost} presents the computation costs of our method compared to prior methods. The extra computation cost of our proposed method to OFA is path sampling and path filter fine-tuning after supernet warm-up (lines 4 and 5 in Algorithm \ref{alg:supernet training}) and path pruning (lines 12 and 13 in Algorithm \ref{alg:supernet training}). The computation cost for path filter fine-tuning was negligible (less than 2 min on one RTX A4000 GPU) compared to other processes. Also, the path pruning cost did not affect the training cost for COCO. It was about a +10\% increase for Pascal VOC. The proposed method outperforms OFA (w/ PS) although the computation cost is reduced by approximately 30\% and 60\% for Pascal VOC and COCO, respectively because our method efficiently trains the supernet without progressive shrinking by pruning the search space. This suggests that our method effectively prunes the search space to be trained with small detection datasets. Furthermore, we confirmed that when OFA and CompOFA* supernets are trained for the same GPU costs as ours, our method still outperforms (detailed in the supplementary materials, Section \ref{sec:evaluation for the same GPU costs}).  Compared to the fullnet performance, the best-performing paths found by the evolutionary search were approximately 0.50 and 1.00 points worse for Pascal VOC and COCO, respectively; the method trains multiple paths well without fine-tuning.

Further, the results of a uniform sampling of 15 paths in each FLOPs bucket are presented in Figure \ref{fig:eval search space}. Paths are sampled from $\mathcal{A}$ and $\Tilde{\mathcal{A}}$ for OFA (w/o PS) and ours, respectively. For both datasets, the proposed method presents a larger population of paths with better performance for overall FLOPs than OFA (w/o PS). This confirms our method's ability to obtain good-performing paths for multiple resource constraints by search space pruning via the path filter.






\subsection{Path filter performance}

Table \ref{tab:path filter performance} summarizes the path filter performance under different settings explained in Section \ref{sec:path filter design} for the weakest 25\% path prediction. The proposed FLOPs-bounded loss, budget conditioning, and block positional encoding improve the precision by 24\% 2\%, and 2\%, respectively.


\subsection{Analysis on path and operation pruning}

\begin{figure}[t]
\begin{center}
   \includegraphics[width=1.0\linewidth]{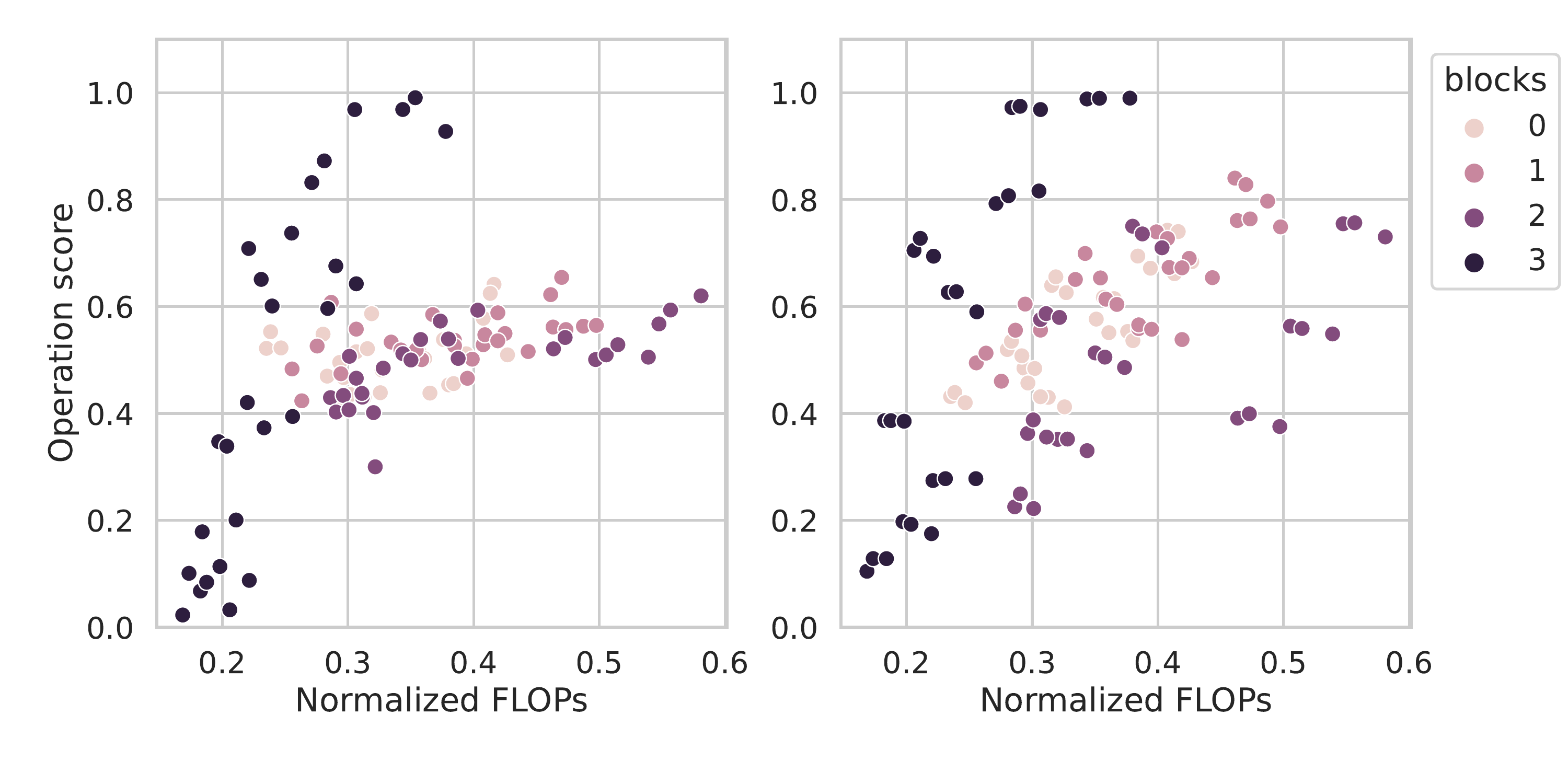}
\end{center}
   \caption{Operation scores for the supernet trained by (left) ImageNet and (right) Pascal VOC. For pascal VOC, the score is more affected by operation configurations such as block id and depth.}
\label{fig:operation scores}
\end{figure}

\begin{figure}[t]
\begin{center}
   \includegraphics[width=1.0\linewidth]{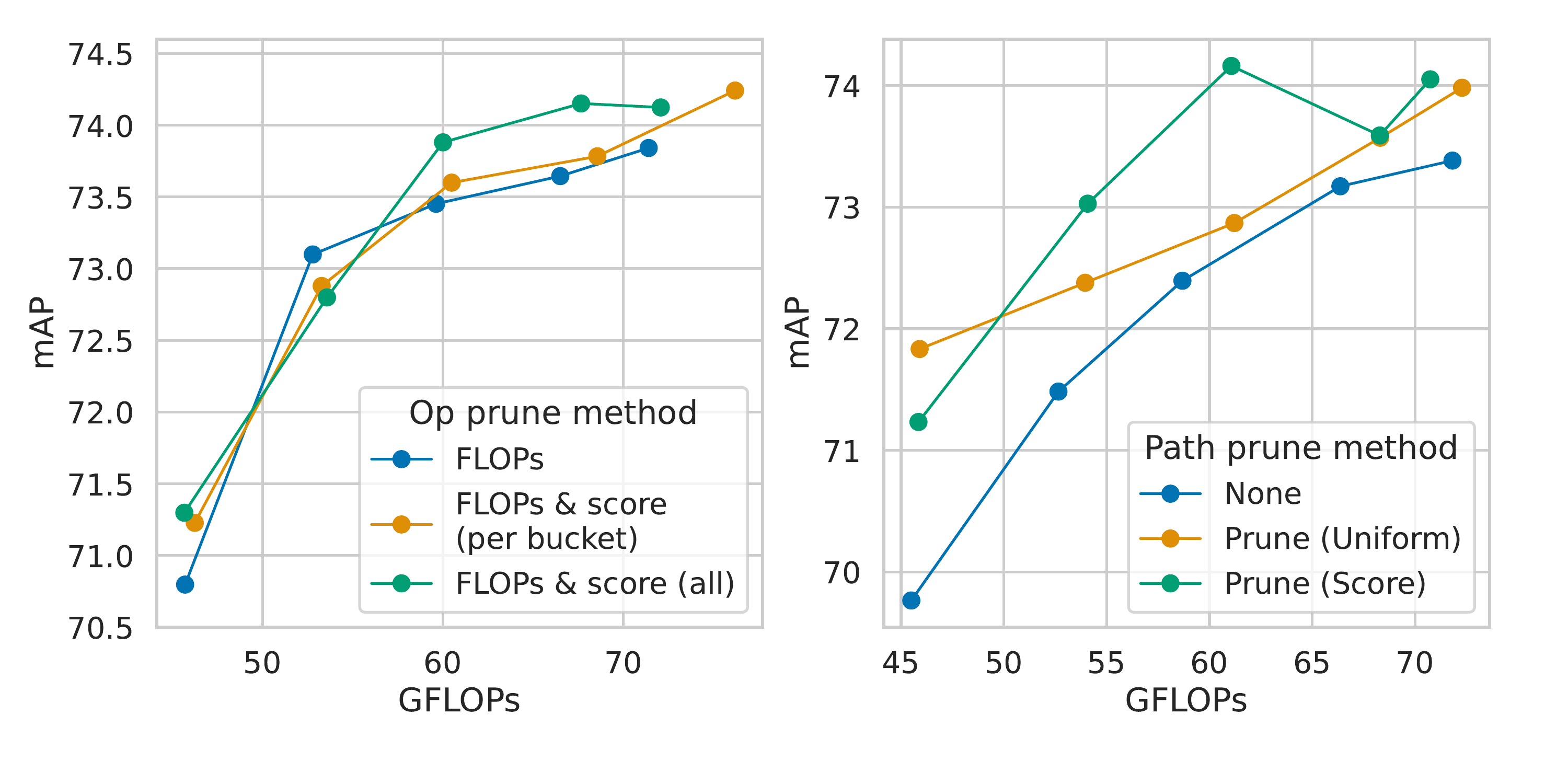}
\end{center}
   \caption{Results for different operation and path pruning methods. For operation pruning (left), FLOPs \& score (all), which removes weak performing operations over all layers, performs the best. For path pruning (right), score-based pruning outperforms uniform sampling.}
\label{fig:results for operation and pruning methods}
\end{figure}

\begin{figure}[t]
\begin{center}
   \includegraphics[width=1.0\linewidth]{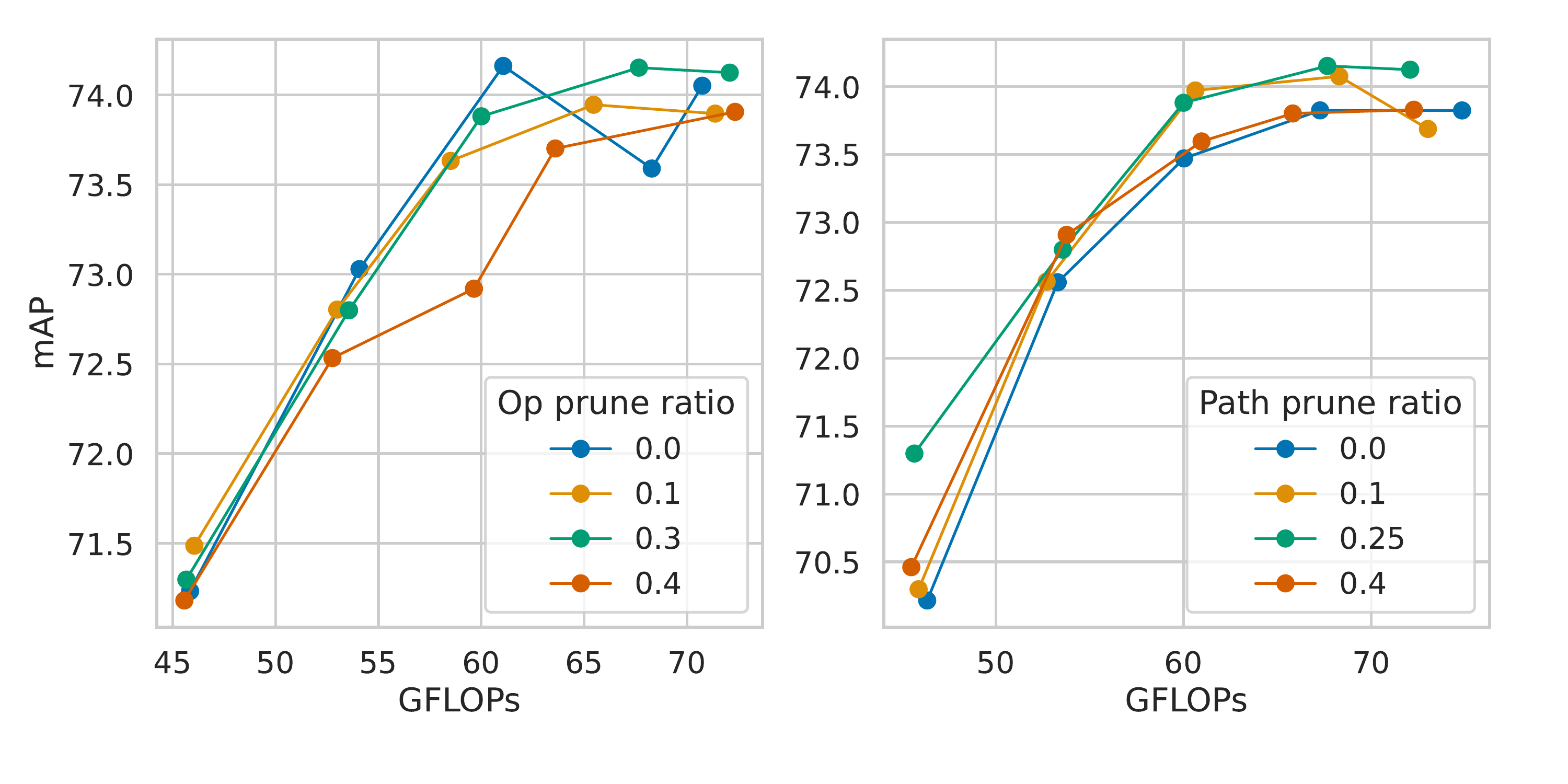}
\end{center}
   \caption{Results for different pruning ratios for (left) operations and (right) paths on Pascal VOC dataset. Without operation nor path pruning ($r_{\mathrm{path}}, r_{\mathrm{op}}=0$), the performance degrades.}
\label{fig:performance on different pruning ratios}
\end{figure}

Figure \ref{fig:operation scores} presents the operation scores calculated by the method explained in Section \ref{sec:search space pruning} for the supernet trained by ImageNet (after pre-training) and Pascal VOC (after supernet warm-up). Compared to the pre-trained supernet, the fine-tuned supernet presents a stronger correlation between operation scores and configuration such that within each block, the larger the operation size, the larger the operation score. This implies that detection models' performance differs more depending on its architecture than classification; it motivates our operation pruning strategy of removing weak ones because pruning operations without considering performance might remove good ones.

The effectiveness of score-based search space pruning is analyzed in Figure \ref{fig:results for operation and pruning methods}. For both, only operation or path pruning was performed. The score-based pruning outperforms other methods. For operation (left in \ref{fig:results for operation and pruning methods}), "FLOPs \& score (all)" performs the best. This highlights the effectiveness of removing weak-performing operations. For path (right in \ref{fig:results for operation and pruning methods}), the score-based method performs well except for the smallest FLOPs budget. This may be due to the lack of operation pruning; removing weak-performing operations is especially important to train smaller paths. The performances across different pruning ratios are evaluated (Figure \ref{fig:performance on different pruning ratios}). Without path pruning (that is, $r_{\mathrm{path}} = 0$) performs worse for all FLOPs-bounds. Without operation pruning performs the worst for the smallest FLOPs. This suggests that operation pruning is more effective for small paths where performance degradation is more obvious without training tricks. Too much pruning (for example, $r = 0.4$) also performs weaker than other ratios; this may be due to removing some good-performing candidate operations or paths.


\section{Conclusion}
In this paper, we propose a search space pruning technique for training a supernet. The search space is pruned by using a path filter that learns the relative performance of paths. More specifically, based on the given resource constraints and on the predicted path rankings, weak operations and paths are removed from the search space. The experimental results show that our proposed method yields better-performing supernets for object detection. Future works include extending our path filter to more complex search spaces and further optimizing architectures of up-convolutional and head structures in addition to the backbone. Our proposed method could also be extended to other vision tasks such as segmentation tasks and human pose estimation.


\section*{Acknowledgement}
\noindent
We would like to thank Lukas Mauch for revising this paper.


{\small
\bibliographystyle{ieee_fullname}
\bibliography{egbib}
}


\clearpage

\appendix
\begin{appendices}

\setcounter{table}{0}
\renewcommand{\thetable}{A\arabic{table}}

\setcounter{figure}{0}
\renewcommand{\thefigure}{A\arabic{figure}}


\section{Related works on object detection} \label{sec:related works on object detection}

Recently, two-stage detectors \cite{girshick2015fast, ren2015faster} have been performance state-of-the-art.
They first generate class-independent region proposals by using the region proposal network, then classify them by using detection heads.
However, they have the drawbacks of long inference time and complex model architecture.
To cope with this drawback, one-stage detectors \cite{redmon2016you, lin2017focal} directly predict object categories and bounding boxes (that is, anchors) at each location of feature maps that are generated by the backbone network.
Although this end-to-end approach has the advantage of faster inference, it  requires hyper-parameter tuning to find suitable anchors and complex model architecture for increasing the number of anchors.


\section{Training details} \label{sec:training details}

\begin{table}[h]
\begin{center}
\begin{tabular}{|l|l|}
\hline
Stage & Search space, $($D$, $W$, $E$)$ \\
\hline\hline
1     &  $\{0, 1, 2\} \times \{1.0\} \times \{0.35\}$  \\
\hline
2     &  $\{0, 1, 2\} \times \{0.8, 1.0\} \times \{0.35\}$  \\
\hline
3     &  $\{0, 1, 2\} \times \{0.65, 0.8, 1.0\} \times \{0.35\}$  \\
\hline
4     &  $\{0, 1, 2\} \times \{0.65, 0.8, 1.0\} \times \{0.25, 0.35\}$  \\
\hline
5     &  $\{0, 1, 2\} \times \{0.65, 0.8, 1.0\} \times \{0.2, 0.25, 0.35\}$  \\
\hline
\end{tabular}
\end{center}
\caption{Search space for OFA PS.}
\label{tab:progressive shrinking search space}
\end{table}

\begin{table}[h]
\begin{center}
\begin{tabular}{|l||c|c|c|c|c|c|}
\hline
Stage  & 1  & 2 & 3 &  4 & 5\\
\hline
Epochs & 70 & 5  & 65 & 5 & 65 \\
\hline
\end{tabular}
\end{center}
\caption{Training schedule for OFA progressive shrinking.}
\label{tab:progressive shrinking training schedule}
\end{table}

The search space and training schedule for each OFA progressive shrinking (PS) stage are detailed in Table \ref{tab:progressive shrinking search space} and Table \ref{tab:progressive shrinking training schedule}.
For training, we used the Adam optimizer \cite{kingma2014adam}.
Settings for each dataset are detailed as follows.

\textbf{Pascal VOC}:
The initial learning rate was set to 5e-4, with the step scheduler for learning rate decay.
The learning rate was decayed by 0.1 at 45 and 60 epochs.
The training epochs for fullnet were 70.
The training batch size was 32.

\textbf{COCO}:
The initial learning rate was set to 5e-4, with the cosine scheduler for learning rate decay.
The fullnet training epochs were 140.
The training batch size was 64.


\section{Evalution of path filter} \label{sec:path filter evaluation}


\begin{table}[h]
\begin{center}
\begin{tabular}{|l||c|c|c|c|c|}
\hline\
Pruning ratio  & Accuracy & Precision & Recall \\
\hline\hline
0.2            & 0.940    & 0.850     & 0.180 \\
\hline
0.3            & 0.910    & 0.869     & 0.252 \\
\hline
0.4            & 0.890    & 0.864     & 0.368 \\
\hline
\end{tabular}
\end{center}
\caption{Path filter performance for predicting the weakest $r_{\mathrm{path}}$\% paths on different pruning ratios. Our path filter can be used for different pruning ratios.}
\label{tab:path filter performance on different pruning ratios}
\end{table}

Our path filter is designed to predict the relative performance of paths.
It is more flexible than the path filter proposed in the prior work \cite{huang2022greedynasv2}, i.e., once the path filter is trained, a different pruning ratio can be applied.
Here, we demonstrate that our path filter performs well when different pruning ratios are adopted.
Table \ref{tab:path filter performance on different pruning ratios} presents the path filter performance to predict the weakest $r_{\mathrm{path}}$\% paths.
For all pruning ratios, the precision is more than 0.7.
The performance, especially for precision and recall, improves for larger pruning ratios because classification is easier when the number of positive and negative samples is similar.
The results confirm the utility of our path filter for different pruning ratios.

\section{Comparision with prior NAS approaches under the same GPU costs} \label{sec:evaluation for the same GPU costs}

\begin{figure}[t]
\begin{center}
   \includegraphics[width=1.0\linewidth]{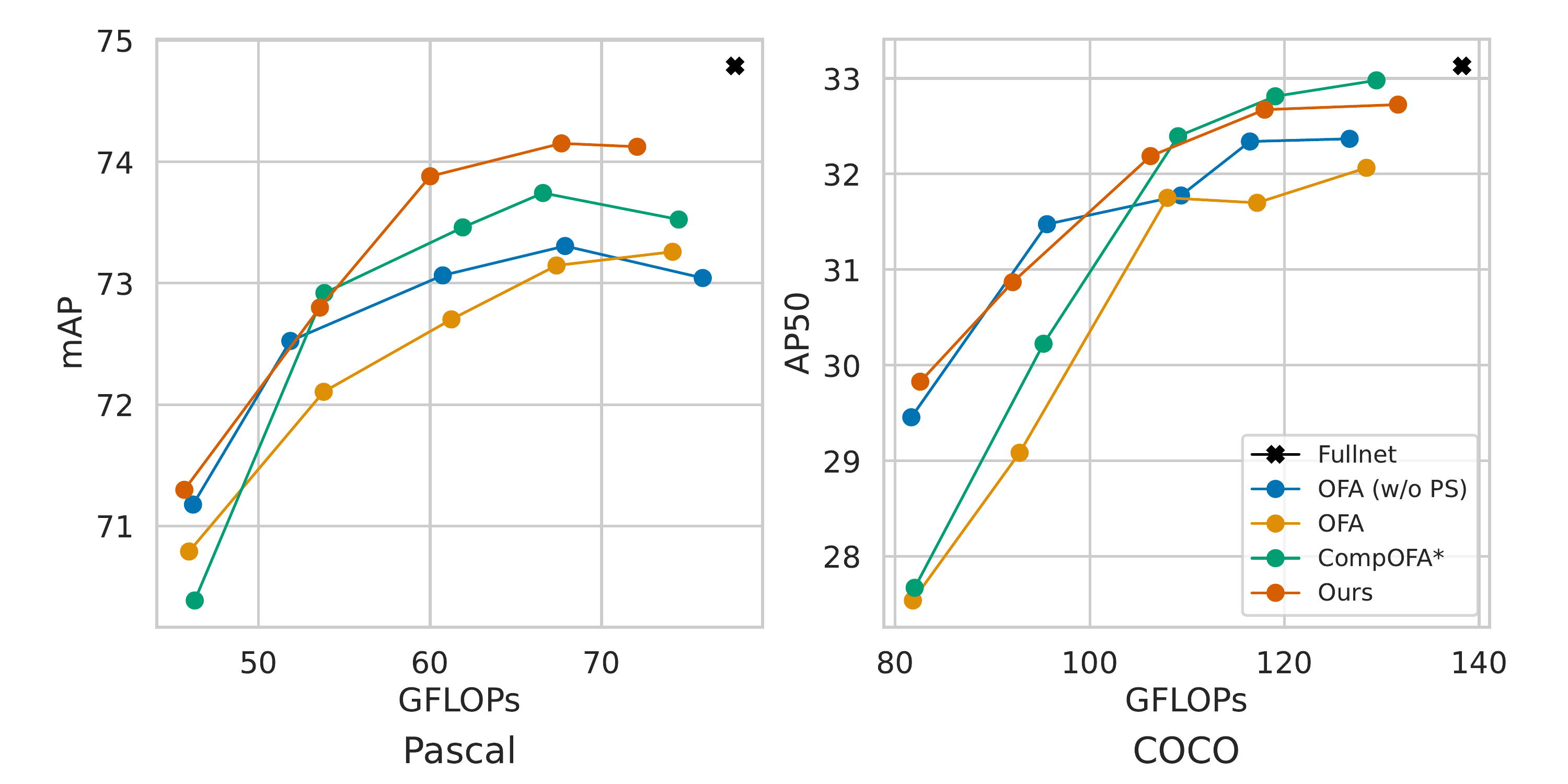}
\end{center}
   \caption{The performance of the optimal architecture $a^*$, for object
detection on the Pascal VOC (left) and the COCO (right)
dataset. Our method outperforms OFA (w/o progressive shrinking,
PS) across all given FLOP bounds. CompOFA* performs well
only for larger FLOPs.}
\label{fig:evolution search results for the same GPU costs}
\end{figure}

\begin{table}[h]
\begin{center}
\begin{tabular}{|l||c|c|}
\hline\
Method  & 
\begin{tabular}{c} Epochs \\ (Pascal VOC) \end{tabular} &
\begin{tabular}{c} Epochs \\ (COCO) \end{tabular} \\
\hline\hline
OFA (w/o PS) & 147 & 147 \\
\hline
CompOFA*     & 147 & 147 \\
\hline
Ours         & 70  & 140 \\
\hline
\end{tabular}
\end{center}
\caption{Training schedule for comparison under the same GPU costs.}
\label{tab:training schedule for comparison under the same GPU costs}
\end{table}

Figure \ref{fig:evolution search results for the same GPU costs} presents the evolution search results for training the supernet with OFA (w/o PS) and CompOFA* for the same GPU costs with the proposed method. The training schedule is summarized in Table \ref{tab:training schedule for comparison under the same GPU costs}. OFA (w/ PS) is trained for the same schedule as results in Figure \ref{fig:evolution search} and presented for reference. For both Pascal VOC and COCO, training more epochs improves the accuracy of small paths for OFA, however, it degrades the accuracy for large paths. Moreover, the accuracy of small paths for CompOFA* is smaller than that of ours. This infers the limitation of hand-crafted search space pruning. 


\end{appendices}

\clearpage



\end{document}